\documentclass{article}
\usepackage{stywhispers,amsmath,epsfig}
\usepackage{graphicx}
\usepackage{color,soul}
\usepackage{hyperref}
\usepackage{mwe}
\usepackage{subcaption}
\usepackage{amsmath}
\usepackage{todonotes}
\usepackage{tabularx} % in the preamble
\usepackage{multirow}
\usepackage{booktabs}

\newcolumntype{L}{>{\centering\arraybackslash}m{3cm}}
\usepackage{float}
\usepackage[thinc]{esdiff}
\usepackage{graphicx} % omit 'demo' for real document
\usepackage[T1]{fontenc}
\usepackage{caption}
\usepackage{enumitem}
\usepackage{listings}
\usepackage[htt]{hyphenat}

\setlist[itemize]{noitemsep, topsep=0pt}
\newlength{\bibitemsep}
\newlength{\bibparskip}
\let\oldthebibliography\thebibliography
\renewcommand\thebibliography[1]{%
  \oldthebibliography{#1}%
  \setlength{\parskip}{\bibitemsep}%
  \setlength{\itemsep}{\bibparskip}%
}
\def\cuvisai{\textit{cuvis.ai}}

\lstdefinestyle{cuvisaistyle}{
  language=Python,
  xleftmargin=2em,
  xrightmargin=0.3em,
  frame=single,
  framexleftmargin=2em,
  stringstyle=\color{darkgreen},
  keywordstyle=\color{blue},
  commentstyle=\color{grey},
  numbers=left,
  stepnumber=1,
  numbersep=0.7em,
  tabsize=2,
  showspaces=false,
  showstringspaces=false,
  basicstyle=\footnotesize\ttfamily,
  backgroundcolor=\color{white},
  numberstyle=\ttfamily\color{lightgrey}
}

\definecolor{darkgreen}{rgb}{0.0,0.5,0.1}
\definecolor{grey}{gray}{0.3}
\definecolor{codebg}{rgb}{0.95,0.95,0.92}
\definecolor{lightgrey}{gray}{0.65}

\usepackage{microtype}

\title{Cuvis.Ai: An Open-Source, Low-Code Software Ecosystem for Hyperspectral Processing and Classification}
\name{\scalebox{0.96}{Nathaniel Hanson$^{1,2}$, Philip Manke$^{2}$, Simon Birkholz$^{2}$, Maximilian Mühlbauer$^{2}$, Rene Heine$^{2*}$, Arnd Brandes$^{2}$
\thanks{*Corresponding author: {\tt\small heine@cubert-gmbh.de}}}}
\address{$^{1}$Institute for Experiential Robotics; Northeastern University, Boston, Massachusetts, USA\\$^{2}$Cubert GmbH, Ulm, Germany}
\begin{document}
%\ninept
%
\maketitle
\begin{abstract}
Machine learning is an important tool for analyzing high-dimension hyperspectral data; however, existing software solutions are either closed-source or inextensible research products.
In this paper, we present \cuvisai{}, an open-source and low-code software ecosystem for data acquisition, preprocessing, and model training.
The package is written in Python and provides wrappers around common machine learning libraries, allowing both classical and deep learning models to be trained on hyperspectral data.
The codebase abstracts processing interconnections and data dependencies between operations to minimize code complexity for users.
This software package instantiates nodes in a directed acyclic graph to handle all stages of a machine learning ecosystem, from data acquisition, including live or static data sources, to final class assignment or property prediction.
User-created models contain convenient serialization methods to ensure portability and increase sharing within the research community.
All code and data are available online: \url{https://github.com/cubert-hyperspectral/cuvis.ai}
\end{abstract}

\section{Introduction}
\label{sec:intro}
Hyperspectral imaging (HSI) is growing in importance as a sensory technique in a variety of domains including healthcare \cite{karim2023hyperspectral}, precision agriculture \cite{khan2018modern}, mining \cite{dieters2024robot}, and robotics \cite{hanson2022occluded,hanson2023hyperdrive}.
Significantly, HSI has been enabled by advances in sensor technology that allow smaller, lighter, and more cost-effective means of observing the electromagnetic spectrum with spatial information \cite{west2018commercial}.
As the number of sensor manufacturers increases, so does the volume of research focused on HSI classification.
Although the academic literature is replete with fundamental advances in deep learning methods to build classifiers, regressors, and anomaly detectors \cite{li2019deep}, research struggles to adapt to a general audience who are not intimately familiar with machine learning (ML) libraries, graphical processing unit (GPU) acceleration, and model architecture selection.
The commercial solutions for HSI ML model training that exist are closed-source and do not accommodate new model types.

\begin{figure}[t]
    \includegraphics[width=\linewidth]{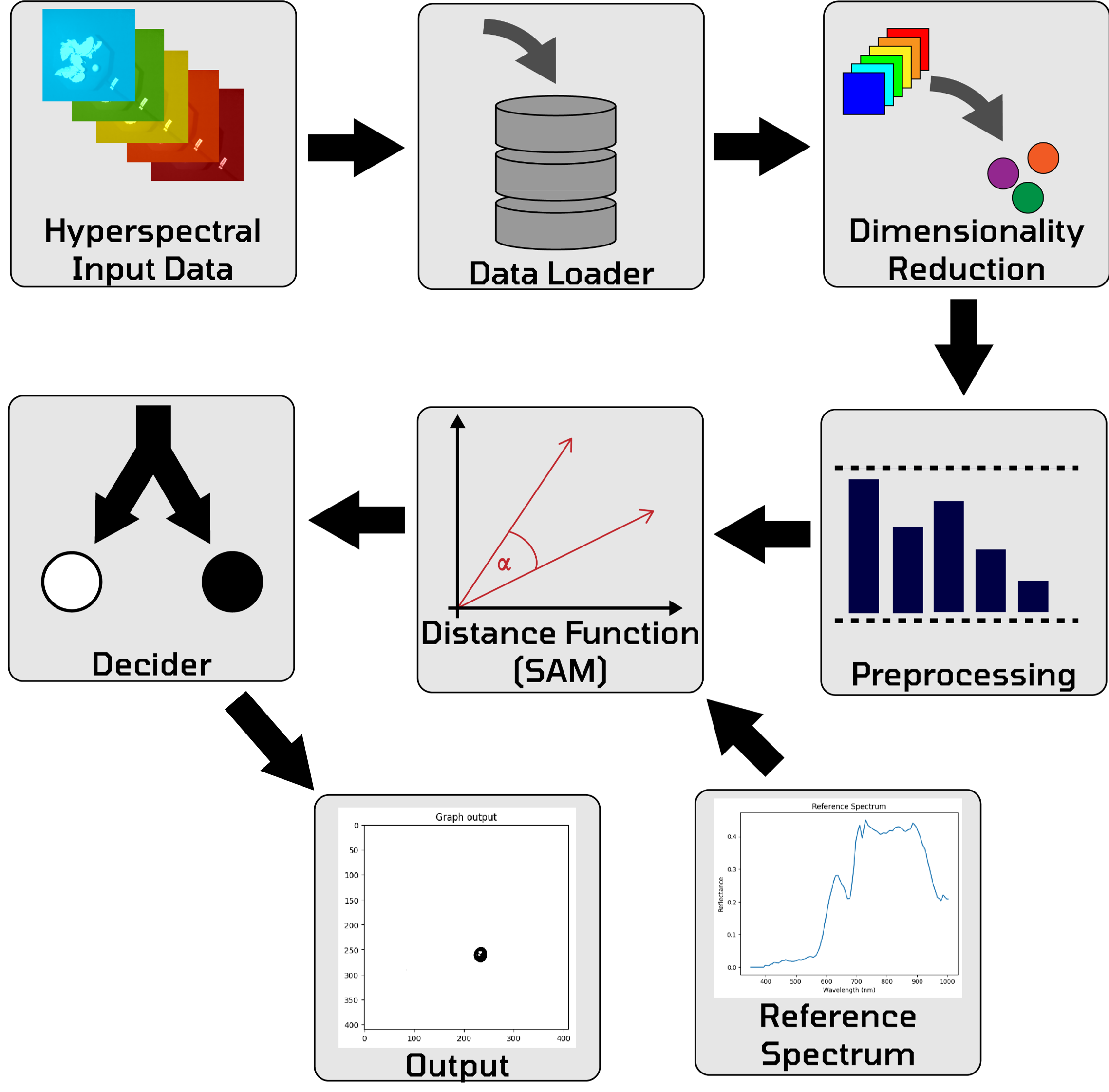}
    \caption{Exemplar processing graph within the \textit{cuvis.ai} framework containing a data loader, dimensionality reduction, preprocessing, distance comparisons to reference spectra, and decision function.}
    \label{fig:cuvis_ai_teaser}
    \vspace{-2.0em}
\end{figure}

To address this shortcoming in the literature, we introduce \cuvisai{}, a free, open-source, extensible library for HSI model training and analysis.
This library bridges the experience gap between data practitioners and ML scientists, by providing a low-code interface to build ML pipelines that transform HSI data into decision points for end users.
The code base provides intuitive abstractions that enable users to contribute new models and algorithms for the benefit of the entire HSI research and user communities.

Fig.~\ref{fig:cuvis_ai_teaser} illustrates an example processing graph created using \cuvisai{}, containing data loading, data transformation, and decision stages.
Specifically, \cuvisai{} advances the state-of-the-art (SOTA) in HSI software through:
\begin{itemize}
    \item Open-source software framework for hyperspectral data processing and analysis.
    \item Graph-based organization of hyperspectral imaging model training and inference.
    \item Template for contributing new models to integrate research models into the code base.
\end{itemize}

\section{Related Work}
\label{sec:prior}

In recent years, researchers have developed tools and collected datasets to enable scientists to explore HSI data through ML.
In this section, we review relevant academic and commercial advances to contextualize \cuvisai{}'s contributions.

\subsection{Languages \& Machine Learning Packages}

In the field of machine learning, \textit{Python} is the standard programming language \cite{innes2018machine}.
It is a dynamically typed language with highly readable syntax that provides a low barrier to entry.
While other languages are also used in HSI research, for example \textit{R}~\cite{lehnert2018hyperspectral} or \textit{MATLAB}~\cite{maurya2018hicf,mobaraki2018hyper}, they lack a similar open-source ecosystem that has been built around \textit{Python}.
This ecosystem includes packages such as \textit{scikit-learn}~\cite{pedregosa2011scikit} for more classical algorithms and \textit{PyTorch}~\cite{paszke2019pytorch} for deep learning applications.
While these powerful frameworks support complex and cutting-edge research, full and accurate use of their capabilities requires expert knowledge in the field of software development; thus, their general uptake by users with primary expertise in HSI application-driven projects is limited.

\subsection{Hyperspectral Packages}
\cite{mishra2021idcube} proposes an open-source graphical user interface (GUI) for HSI data processing.
This solution focuses on processing algorithms and statistical techniques for HSI data analysis.

\cite{SOFTWARE_PyHat} is another open-source software tool for HSI processing and inference.
It provides both a GUI and an importable \textit{Python} package for visualization and statistical analysis.
Similarly, deep learning models are not supported.

\cite{SOFTWARE_SpectralPython,annala2018practical} provide interfaces for algorithms commonly used in analyzing HSI data, but focus on classical machine learning algorithms.
These packages use optimized libraries with C-language bindings, such as \textit{numpy} \cite{harris2020array} to expedite large-array calculations.
Since these packages are designed primarily for exploratory analysis, they do not provide a utility to deploy learned ML pipelines with live camera data.

% \cite{annala2018practical} present a pipeline for developing HSI analysis tools.
% They leverage numpy \cite{harris2020array}, pandas \cite{mckinney-proc-scipy-2010} and scikit-learn for this endeavor, focusing on modularity and openess to develop new features.
% While they provide code examples and tools for combining the employed frameworks, they do not provide an end-to-end solution allowing a user to easily construct an entire processing pipeline with little code.

\cite{SOFTWARE_HSIToolbox} developed a web-based graphical interface to guide domain experts through the process of classifying hyperspectral images, from uploading the image to a deep learning model ready for training.
They focus on the users tuning parameters of a few pre-selected algorithms and curating labeled data.
Like other solutions, this technology lacks a generic interface to novel models.
Other packages follow a strategy of compiling multiple algorithms~\cite{audebert2019deep} into a single installable package, but lack regular software support and updates.

\subsection{Commercial Tools}

Manufacturers have developed commercially available toolkits for HSI ML but their products lack the same open-contribution model of general ML packages that advanced the SOTA of research and applications.
For example, Prediktera (Umeå, Sweden), SpecimOne (Oulu, Finland), and perClass Mira (Delft, Netherlands) are closed source and are not extensible to new model types.
Some companies, such as EVK and Insort (Austria), have completely closed ecosystems that require the use of their software with their hardware.
More open approaches also exist, such as HALCON by MVTec (Munich, Germany), but these solutions focus on generic image processing and are not specialized for HSI.

\section{Software Design}
\label{sec:system}

Considering the identified gaps in the available software for HSI analysis and model building, we design \cuvisai\ to provide versatility and ease of use to users of all levels of experience.
A \textit{cuvis.ai} ML system consists of three main parts:
\begin{itemize}
\item {\textbf{Dataset:} The data provider for the system. Currently, a loader for data in \textit{numpy}~\cite{harris2020array} and typical image formats, and loaders for live data from Cubert and IMEC snapshot cameras are available.}
\item {\textbf{Nodes:} Stages in the system which apply transformations on the data that are passed to it. It is an abstract type with concrete implementations given the intended operation of the node.}
\item {\textbf{Graph:} The manager of interconnections between nodes. The graph manages the relationships and requirements for the dimensionality and data sources and orchestrates the execution of training and inference.}
\end{itemize}

\begin{figure}

\begin{lstlisting}[style=cuvisaistyle]
import cuvis_ai

# Define PCA with 10 components
pca = cuvis_ai.preprocessor.PCA(10)

# Define a K-Means unsupervised classifier
kmeans = cuvis_ai.unsupervised.KMeans(2)

# Define binary decider with threshold value
decider = cuvis_ai.deciders.BinaryDecider()

# Define and construct graph
graph = cuvis_ai.pipeline.Graph("DemoGraph")
graph.add_base_node(pca)
graph.add_edge(pca, kmeans)
graph.add_edge(kmeans, decider)

# Define unlabeled dataset
data = cuvis_ai.data.NumpyDataSet("./demodata")

# Fit graph using first data element
graph.fit(*data[0])

# Run inference with all data
result = graph.forward(*data)
\end{lstlisting}
\caption{Simple example of constructing and using \cuvisai\ to preprocess and classify HSI data with minimal lines of code.}
\vspace{-2.0em}
\label{fig:simple_code_example}
\end{figure}
Fig.~\ref{fig:simple_code_example} shows a minimal code example that imports, constructs, fits and runs a simple ML system using \cuvisai{}.
To describe a data processing system, first the nodes that will comprise it are defined and configured, as shown in lines 3-10.
Lines 13-16 instantiate the graph and add and describe the nodes' relations to each other using the \lstinline[style=cuvisaistyle]{add_edge()} method. 
Next, the graph is fit using data from the dataset created in line 19, by passing the first data element into the \lstinline[style=cuvisaistyle]{fit()} method in line 22.
Finally, new data are processed as shown in line 25 by passing the second data element of the dataset through the graph.
The complete data loading, preprocessing, and classification procedure requires only 11 lines of code and no intermediary data products to be passed between functions.

\subsection{Data Representation}

\textit{Cuvis.ai} ensures compatibility with the algorithms of the constituent frameworks by utilizing \textit{numpy} arrays internally to pass the HSI data through the graph.
This array representation allows for integration of algorithms from ML frameworks, such as \textit{scikit-learn}~\cite{pedregosa2011scikit}, \textit{PyTorch}~\cite{paszke2019pytorch}.

The framework includes dataloaders for generic data, such as \textit{numpy} arrays or TIFF files, as well as specialized dataloaders for already recorded data from Cubert cameras and for live inference from connected snapshot cameras.
These dataloaders include a preprocessing stage, leveraging the \textit{Torchvision} package from \textit{PyTorch}.
This stage allows for on-the-fly preprocessing, such as resizing, normalization, and band selection, for both the data cubes and labels. 

To further improve versatility, \cuvisai\ enriches the data used in the graph.
Along with hyperspectral data cubes, labels and meta-data are also passed through the graph.
These additional data elements are only utilized by nodes that request them, minimizing the performance impact of this approach.

The label data are in the standardized \textit{COCO}~\cite{lin2014microsoft} JSON format, to ensure compatibility, containing object classifications, segmentation masks, bounding boxes, and more. 
The meta-data are stored in a \textit{Python} dictionary with clearly named data fields and persisted in YAML format containing original file names and cube dimension sizes, wavelength to channel mappings, reference cubes, and more.

Labels and meta-data are passed through the graph during training and inference, providing additional channels for information to be passed between nodes without modifying the HSI data.
The vision for this universal side-channel is to provide a way for nodes to communicate intermediate results between each other. This allows the design of multi-step algorithms as a single larger graph, without having to rely on larger deep learning networks or having to build interfaces between separate graphs.
For example, a graph designed and trained for object detection of apples might need to be extended into a regression system to detect the ripeness level of segmented apples.
Using \cuvisai{}, the apple detection graph can be easily expanded for this task by feeding the generated detection mask labels directly to the nodes added and trained for ripeness estimation.

\subsection{Design Decisions}
\label{ssec_design_desicions}

In \cuvisai\ different ML algorithms are categorized by their use-cases.
The node's category informs the graph of which data (numeric, labels, metadata) the node expects.
This approach is also useful from a user perspective as it provides a pre-sorted collection of algorithms:

\begin{itemize}
\item \texttt{Transformation:} Nodes that execute simple mathematical transformations on the data and potentially on labels, such as normalization, resizing and cropping.
\item \texttt{Preprocessor:} Nodes that affect the spectral dimension of the data, such as Principal Components Analysis (PCA) or Non-Negative Matrix Factorization (NMF).
\item \texttt{Distance:} Nodes comparing spectra to a reference to provide a similarity score, such as Euclidean distance or Spectral Angle Mapper (SAM) and others from \cite{deborah2015comprehensive}.
\item \texttt{Unsupervised:} Nodes trained on some example data but do not require labels, such as K-Means Clustering or Gaussian Mixture Model (GMM).
\item \texttt{Supervised:} Nodes that are trained on labeled data, such as Support Vector Machine (SVM), Quadratic Discriminant Analysis (QDA) or deep learning models.
\item \texttt{Deciders:} Nodes transform outputs of supervised nodes with class predictions into decisions (i.e. labels) using, for example, a simple threshold or a maximum likelihood selection.
\end{itemize}

Following object-oriented design principles, the minimalist base class for nodes -- comprising methods for data processing, providing input and output dimensionality, serialization, and loading -- enables the development of custom nodes with ease. Each node also stores expected input and output dimensions.
This allows the graph to perform a verification step by making sure that the graph contains congruous dimensions between processing nodes.

The acyclic graph is employed as the topology of the system to allow for branching and merging of the data flow.
This enables intermediate results, the hyperspectral data, labels or meta-data, to be easily re-used multiple times in other stages of the graph.
The execution model of \cuvisai\ takes the lifetime of the data and removes intermediary products once all nodes that require it as input have computed their products. Graph training and inference are achieved by first performing a topological sort on the graph and executing the nodes in the resulting optimal order.

\subsection{Serialization \& Distribution}

\textit{Cuvis.ai} is envisioned as an open tool to further the research efforts of the HSI community.
To support this vision, the framework contains methods for saving and restoring processing graphs.
Any \cuvisai\ system can be stored on disk as a compressed zip directory, comprised of the internal states of all nodes stored as 
\texttt{.npl} or \texttt{.zip} files along with a main YAML file that describes the topology of the graph and the nodes' attributes in a human readable format.
This implementation also contains versioning to detect incompatible versions of nodes.
Furthermore, the framework provides a tool for downloading openly available datasets compatible with \cuvisai{}.

\subsection{Code Structure \& Contributing}

\textit{Cuvis.ai}'s code is structured into submodules matching key areas of the framework.
These submodules include:
\texttt{data}: containing the datasets and utilities;
\texttt{node}: containing the node base class.
\texttt{pipeline}: containing the graph class.
Most other submodules are named after the categories of the nodes that implement different algorithms and processing steps, as mentioned in Section~\ref{ssec_design_desicions}.

The code is formatted according to PEP8 and further stylistic choices are explained in the project's \texttt{CONTRIBUTING.md} file.
\cuvisai\ is hosted on GitHub and is fully open source under the Apache v2\footnote{\url{https://www.apache.org/licenses/LICENSE-2.0}} software license.
Contributions to \cuvisai\ are tested using a suite of unit tests that run with every pull request to the project, ensuring code quality and compatibility between versions.
The project is envisioned as an open source, transparent, and expandable alternative to existing frameworks for hyperspectral data analysis.
The use of, and contributions to \cuvisai\ are welcome and encouraged.

\section{DEMONSTRATION}

The processing graph shown in Fig.~\ref{fig:cuvis_ai_teaser} is also used in this section to demonstrate the use of \cuvisai{} by applying it to an HSI measurement of a tomato and tomato leaves.
The measurement used in this example is publicly available in the "GrowRipe" dataset, which is published within \cuvisai{}.
This hyperspectral image was collected with a Cubert Ultris X20 camera and consists of 164 bands between 350 nm and 1002 nm with a spatial resolution of 410 by 410 pixels.

Four nodes comprise the processing graph, in order: a PCA node reducing the channels to 16, a transformation node applying a normalization, a distance node using SAM, and a binary decision node using a threshold.
The spectrum of the tomato, sampled from a single pixel, is used as the reference spectrum for the distance node with the threshold of the decider node set to a value of 0.25.

\begin{figure}[t]
    \includegraphics[width=\linewidth]{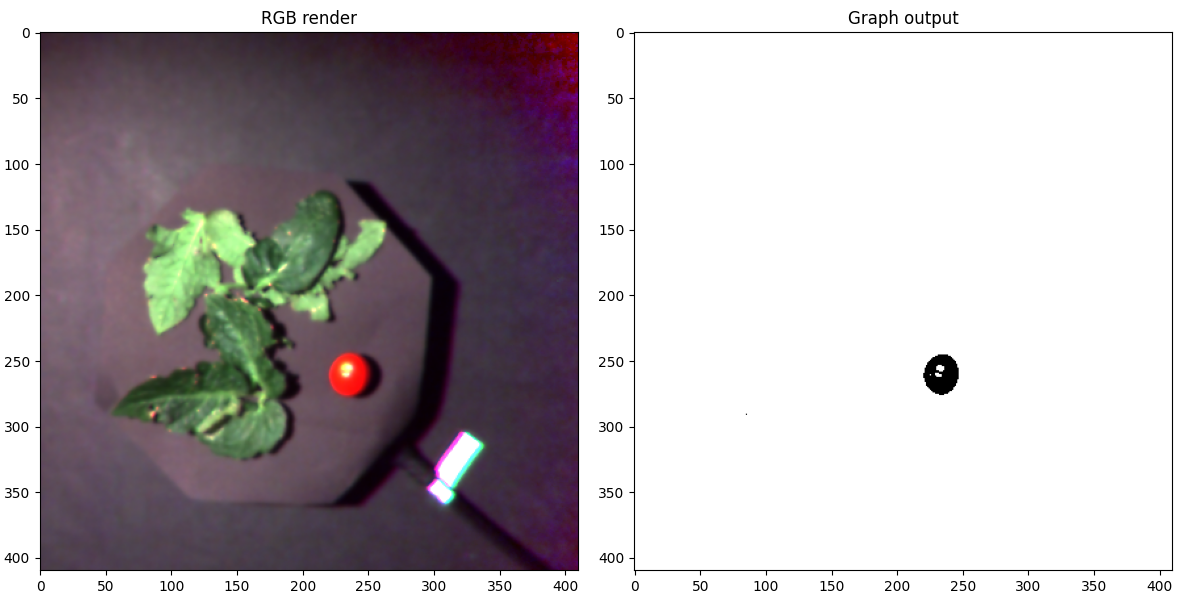}
    \caption{(Left): RGB representation of the input data; (Right): Output of the demo graph using Spectral Angle Mapper (SAM).}
    \label{fig:graph_output}
    % \vspace{-2.0em}
\end{figure}

%\begin{figure}[t]
%    \includegraphics[width=\linewidth]{images/output_KMeans_TomatoNet.png}
%    \caption{(Left): RGB representation of the input data; (Right): Output of the demo graph using KMeans clustering with 5 clusters.}
%    \label{fig:graph_output_kmeans}
%    \vspace{-1.0em}
%\end{figure}

Fig.~\ref{fig:graph_output} shows an RGB rendering of the input data and the resulting image after applying the graph described above to the input.
The system clearly distinguishes between the tomato, background, and leaves present in the image.
A bright spot on the tomato exhibiting specular reflection is not classified correctly due to sensor saturation.
A second graph is also tested to validate the performance of the framework for a different application.
The KMeans-graph replaces the SAM with a KMeans node clustering to 5 clusters but does not use a decider node.

\begin{table}[]
\centering
\caption{Timing of the different steps required to generate and use the graph shown in Fig.~\ref{fig:graph_output} (SAM-Graph) and a slightly altered graph (KMeans-graph). }
\vspace{-0.5em}
\begin{tabular}{l l l}
\toprule
\textbf{Metric} & SAM-graph & KMeans-graph \\ \midrule
Initialization time (s) & 0.005 & 0.002\\ 
Fit time (s) & 0.937 & 0.728 \\ 
Inference time (s) & 0.162 & 0.143\\ 
Model size on disk (kB) & 13 & 21\\ 
Lines of code & 14 & 11 \\ 
\bottomrule
\end{tabular}%
% \vspace{-1.5em}
\label{tab:timing_table}
\end{table}

Table~\ref{tab:timing_table} shows metrics constructed on two different graph models.
The initialization time encompasses constructing and running the graph verification pass.
The serialized graph has a small size of only a few kilobytes, making it easily shareable.
In total, solely counting executed commands, only 14 or 11 lines of code are necessary to build and run the respective systems.
These examples were run on a notebook using an AMD Ryzen 7 6800H processor without GPU acceleration. 

\section{CONCLUSION}
\label{sec:conclusion}

In this work we presented \cuvisai\ -- a first-of-its-kind software ecosystem purpose-built for hyperspectral image processing.
Our system is open-source and extensible to new ML techniques and cameras.
We place a strong emphasis on logical abstraction of classes and data marshaling so that community members can transition research code into reusable and ultimately deployable models for inference on real data.

The code repository contains the core mechanisms to construct, train, and evaluate processing graphs.
In the near future, we will leverage GPUs to accelerate training and inference times of deep learning models.
Currently, the codebase supports snapshot HSI hardware, but will be expanded to support live inference from pushbroom devices.
We will also compose examples of common processing workflows for classification, end-member extraction, and anomaly detection as useful references and will expand the repository of datasets published alongside \cuvisai{}.
Finally, our future work will also add common hyperspectral benchmark examples from~\cite{theisen2024hs3} to demonstrate the quantitative expressiveness of our framework.

In summary, this work presents a unique opportunity to remove obfuscations and barriers to entry in HSI ML by building a community hub for learning, sharing, and collaborating.


\bibliographystyle{IEEEbib}
\bibliography{strings}

\begin{thebibliography}{10}

\bibitem{karim2023hyperspectral}
Shahid Karim, Akeel Qadir, Umar Farooq, Muhammad Shakir, and Asif~A Laghari,
\newblock ``Hyperspectral imaging: a review and trends towards medical imaging,''
\newblock {\em Current Medical Imaging}, vol. 19, no. 5, pp. 417--427, 2023.

\bibitem{khan2018modern}
Muhammad~Jaleed Khan, Hamid~Saeed Khan, Adeel Yousaf, Khurram Khurshid, and Asad Abbas,
\newblock ``Modern trends in hyperspectral image analysis: A review,''
\newblock {\em Ieee Access}, vol. 6, pp. 14118--14129, 2018.

\bibitem{dieters2024robot}
Sibren Dieters,
\newblock ``Robot-aided hyperspectral imaging for mineral exploration in underground mining environments,''
\newblock M.S. thesis, Aalto University, 2024.

\bibitem{hanson2022occluded}
Nathaniel Hanson, Gary Lvov, and Ta{\c{s}}k{\i}n Padir,
\newblock ``Occluded object detection and exposure in cluttered environments with automated hyperspectral anomaly detection,''
\newblock {\em Frontiers in robotics and AI}, vol. 9, pp. 982131, 2022.

\bibitem{hanson2023hyperdrive}
Nathaniel Hanson, Samuel Hibbard, Benjamin Pyatski, Charles DiMarzio, and Ta{\c{s}}k{\i}n Pad{\i}r,
\newblock ``Hyper-drive: Visible-short wave infrared hyperspectral imaging datasets for robots in unstructured environments,''
\newblock in {\em 2023 13th Workshop on Hyperspectral Imaging and Signal Processing: Evolution in Remote Sensing (WHISPERS)}. IEEE, 2023.

\bibitem{west2018commercial}
Michael West, John Grossmann, and Chris Galvan,
\newblock ``Commercial snapshot spectral imaging: the art of the possible,''
\newblock {\em MITRE: McLean, VA, USA}, 2018.

\bibitem{li2019deep}
Shutao Li, Weiwei Song, Leyuan Fang, Yushi Chen, Pedram Ghamisi, and Jon~Atli Benediktsson,
\newblock ``Deep learning for hyperspectral image classification: An overview,''
\newblock {\em IEEE Transactions on Geoscience and Remote Sensing}, vol. 57, no. 9, pp. 6690--6709, 2019.

\bibitem{innes2018machine}
Mike Innes, Stefan Karpinski, Viral Shah, David Barber, PLEPS Saito~Stenetorp, Tim Besard, James Bradbury, Valentin Churavy, Simon Danisch, Alan Edelman, et~al.,
\newblock ``On machine learning and programming languages,''
\newblock Association for Computing Machinery (ACM), 2018.

\bibitem{lehnert2018hyperspectral}
Lukas~W Lehnert, Hanna Meyer, Wolfgang~A Obermeier, Brenner Silva, Bianca Regeling, and J{\"o}rg Bendix,
\newblock ``Hyperspectral data analysis in r: The hsdar package,''
\newblock {\em arXiv preprint arXiv:1805.05090}, 2018.

\bibitem{maurya2018hicf}
Ajay~K Maurya, Divyesh Varade, and Onkar Dikshit,
\newblock ``Hicf: a matlab package for hyperspectral image classification and fusion for educational learning and research,''
\newblock {\em The International Archives of the Photogrammetry, Remote Sensing and Spatial Information Sciences}, vol. 42, pp. 181--188, 2018.

\bibitem{mobaraki2018hyper}
Nabiollah Mobaraki and Jos{\'e}~Manuel Amigo,
\newblock ``Hyper-tools. a graphical user-friendly interface for hyperspectral image analysis,''
\newblock {\em Chemometrics and Intelligent Laboratory Systems}, vol. 172, pp. 174--187, 2018.

\bibitem{pedregosa2011scikit}
Fabian Pedregosa, Ga{\"e}l Varoquaux, Alexandre Gramfort, Vincent Michel, Bertrand Thirion, Olivier Grisel, Mathieu Blondel, Peter Prettenhofer, Ron Weiss, Vincent Dubourg, et~al.,
\newblock ``Scikit-learn: Machine learning in python,''
\newblock {\em the Journal of machine Learning research}, vol. 12, pp. 2825--2830, 2011.

\bibitem{paszke2019pytorch}
Adam Paszke, Sam Gross, Francisco Massa, Adam Lerer, James Bradbury, Gregory Chanan, Trevor Killeen, Zeming Lin, Natalia Gimelshein, Luca Antiga, et~al.,
\newblock ``Pytorch: An imperative style, high-performance deep learning library,''
\newblock {\em Advances in neural information processing systems}, vol. 32, pp. 8026--8037, 2019.

\bibitem{mishra2021idcube}
Deependra Mishra, Helena Hurbon, John Wang, Steven~T Wang, Tommy Du, Qian Wu, David Kim, Shiva Basir, Qian Cao, Hairong Zhang, et~al.,
\newblock ``Idcube lite--a free interactive discovery cube software for multi and hyperspectral applications,''
\newblock in {\em 2021 11th Workshop on Hyperspectral Imaging and Signal Processing: Evolution in Remote Sensing (WHISPERS)}. IEEE, 2021, pp. 1--5.

\bibitem{SOFTWARE_PyHat}
JR~Laura, Lisa~R Gaddis, RB~Anderson, and IP~Aneece,
\newblock ``Introduction to the python hyperspectral analysis tool (pyhat),''
\newblock in {\em Machine learning for planetary science}, pp. 55--90. Elsevier, 2022.

\bibitem{SOFTWARE_SpectralPython}
Thomas Boggs, Don March, Lewis~John McGibbney, François Magimel, kidpixo, The~Gitter Badger, Rajath Kumar, Kirby Banman, and Gemma Mason,
\newblock ``Spectral python 0.21,'' April 2020.

\bibitem{annala2018practical}
Leevi Annala, Matti Eskelinen, Jyri H{\"a}m{\"a}l{\"a}inen, Aamos Riihinen, and Ilkka P{\"o}l{\"o}nen,
\newblock ``Practical approach for hyperspectral image processing in python,''
\newblock in {\em International Archives of the Photogrammetry, Remote Sensing and Spatial Information Sciences}. International Society for Photogrammetry and Remote Sensing, 2018, vol.~42.

\bibitem{harris2020array}
Charles~R Harris, K~Jarrod Millman, St{\'e}fan~J Van Der~Walt, Ralf Gommers, Pauli Virtanen, David Cournapeau, Eric Wieser, Julian Taylor, Sebastian Berg, Nathaniel~J Smith, et~al.,
\newblock ``Array programming with numpy,''
\newblock {\em Nature}, vol. 585, no. 7825, pp. 357--362, 2020.

\bibitem{SOFTWARE_HSIToolbox}
Zeno Dhaene, Nina {\v{Z}}i{\v{z}}aki{\'c}, Shaoguang Huang, Xian Li, and Aleksandra Pi{\v{z}}urica,
\newblock ``Hsitoolbox: A web-based application for the classification of hyperspectral images,''
\newblock {\em SoftwareX}, vol. 22, pp. 101340, 2023.

\bibitem{audebert2019deep}
Nicolas Audebert, Bertrand Le~Saux, and S{\'e}bastien Lef{\`e}vre,
\newblock ``Deep learning for classification of hyperspectral data: A comparative review,''
\newblock {\em IEEE geoscience and remote sensing magazine}, vol. 7, no. 2, pp. 159--173, 2019.

\bibitem{lin2014microsoft}
Tsung-Yi Lin, Michael Maire, Serge Belongie, James Hays, Pietro Perona, Deva Ramanan, Piotr Doll{\'a}r, and C~Lawrence Zitnick,
\newblock ``Microsoft coco: Common objects in context,''
\newblock in {\em Computer Vision--ECCV 2014: 13th European Conference, Zurich, Switzerland, September 6-12, 2014, Proceedings, Part V 13}. Springer, 2014, pp. 740--755.

\bibitem{deborah2015comprehensive}
Hilda Deborah, No{\"e}l Richard, and Jon~Yngve Hardeberg,
\newblock ``A comprehensive evaluation of spectral distance functions and metrics for hyperspectral image processing,''
\newblock {\em IEEE Journal of Selected Topics in Applied Earth Observations and Remote Sensing}, vol. 8, no. 6, pp. 3224--3234, 2015.

\bibitem{theisen2024hs3}
Nick Theisen, Robin Bartsch, Dietrich Paulus, and Peer Neubert,
\newblock ``Hs3-bench: A benchmark and strong baseline for hyperspectral semantic segmentation in driving scenarios,''
\newblock {\em arXiv preprint arXiv:2409.11205}, 2024.

\end{thebibliography}

\end{document}